%% file: 0_main.tex
\pdfoutput=1
\documentclass[11pt,a4paper]{article}
\usepackage[hyperref]{emnlp2018}
\usepackage{times}
\usepackage{latexsym}

\usepackage{url}

\usepackage{multirow}
\usepackage{amsmath, amsfonts, amssymb, xspace, color,listings,enumitem,graphicx}
\usepackage{natbib}
\usepackage{diagbox}
\usepackage[noend,ruled]{algorithm2e}
\usepackage{subfigure}

\usepackage{tabularx, booktabs}
\newcolumntype{Y}{>{\centering\arraybackslash}X}
\definecolor{amber}{rgb}{1.0, 0.75, 0.0}

\def \our {\mbox{AutoNER}\xspace}

\newcommand{\eg}{e.g.,\xspace}

\newcommand{\smallsection}[1]{\smallskip\noindent\textbf{#1.}}

\aclfinalcopy % Uncomment this line for the final submission

\title{Learning Named Entity Tagger using Domain-Specific Dictionary}

\author{
Jingbo Shang $^{\dag*}\ \ $
Liyuan Liu $^{\dag}\thanks{\enspace Equal contribution.}\ \ $
Xiaotao Gu $^{\dag}\ \ $
Xiang Ren $^{\sharp}\ \ $
Teng Ren $^{\ddag}\ \ $
Jiawei Han $^{\dag}$
\\[0.5ex]
{$^{\dag}$ University of Illinois at Urbana-Champaign, Urbana, IL, USA}\\
{$^{\sharp}$ University of Southern California, Los Angeles, CA, USA}\\
{$^{\ddag}$ CooTek Inc., Shanghai, China}\\
{
\begin{small}
\tt
$^{\dag}$\{shang7, ll2, xiaotao2, hanj\}@illinois.edu $^{\sharp}$xiangren@usc.edu $^{\ddag}$teng.ren@cootek.cn
\end{small}
}
}

\date{}

\begin{document}
\maketitle

\begin{abstract}
    \input{0-abs}
\end{abstract}

\input{1-intro}
\input{3-overview}
\input{4-models}
\input{5-dict}

\input{6-exp}
\input{2-related}
\input{7-con}
\input{8-ack}

\newpage

\bibliography{cited}
\bibliographystyle{acl_natbib_nourl}

\end{document}

%% file: 0-abs.tex
%!Tex Root = 0_main.tex

Recent advances in deep neural models allow us to build reliable named entity recognition (NER) systems without handcrafting features.
However, such methods require large amounts of manually-labeled training data.
There have been efforts on replacing human annotations with distant supervision (in conjunction with external dictionaries), but the generated noisy labels pose significant challenges on learning effective neural models. 
Here we propose two neural models to suit noisy distant supervision from the dictionary.
First, under the traditional sequence labeling framework, we propose a revised fuzzy CRF layer to handle tokens with multiple possible labels.
% upgrade the conventional conditional random field (CRF) layer to a fuzzy CRF layer.
After identifying the nature of noisy labels in distant supervision, we go beyond the traditional framework
and propose a novel, more effective neural model \our with a new \texttt{Tie or Break} scheme.
In addition, we discuss how to refine distant supervision for better NER performance.
Extensive experiments on three benchmark datasets demonstrate that \our achieves the best performance when only using dictionaries with no additional human effort, and delivers competitive results with state-of-the-art supervised benchmarks.

%% file: 1-intro.tex
%!Tex Root = 0_main.tex
\section{Introduction}

Recently, extensive efforts have been made on building reliable named entity recognition (NER) models without handcrafting features~\cite{liu2017empower,ma2016end,lample2016neural}.
However, most existing methods require large amounts of manually annotated sentences for training supervised models (\eg neural sequence models)~\cite{liu2017empower,ma2016end,lample2016neural,finkel2005incorporating}. This is particularly challenging in specific domains, where domain-expert annotation is expensive and/or slow to obtain.

To alleviate human effort, distant supervision has been applied to automatically generate labeled data, and has gained successes in various natural language processing tasks, 
including phrase mining~\cite{shang2017automated}, entity recognition~\cite{ren2015clustype,fries2017swellshark,he2017autoentity}, aspect term extraction~\cite{giannakopoulos2017unsupervised}, and relation extraction~\cite{mintz2009distant}.
Meanwhile, open knowledge bases (or dictionaries) are becoming increasingly popular, such as WikiData and YAGO in the general domain, as well as MeSH and CTD in the biomedical domain.
The existence of such dictionaries makes it possible to generate training data for NER at a large scale without additional human effort.

Existing distantly supervised NER models usually tackle the entity span detection problem by heuristic matching rules, such as POS tag-based regular expressions~\cite{ren2015clustype,fries2017swellshark} and exact string matching~\cite{giannakopoulos2017unsupervised,he2017autoentity}.
In these models, every unmatched token will be tagged as non-entity.
However, as most existing dictionaries have limited coverage on entities, simply ignoring unmatched tokens may introduce false-negative labels (e.g., ``prostaglandin synthesis'' in Fig.~\ref{fig:fcrf}).
Therefore, we propose to extract high-quality out-of-dictionary phrases from the corpus, and mark them as potential entities with a special ``unknown'' type.
Moreover, every entity span in a sentence can be tagged with multiple types, since two entities of different types may share the same surface name in the dictionary.
To address these challenges, we propose and compare two neural architectures with customized tagging schemes.

% Firstly, we try to follow the traditional sequence labeling framework.
We start with adjusting models under the traditional sequence labeling framework.
Typically, NER models are built upon conditional random fields (CRF) with the \texttt{IOB} or \texttt{IOBES} tagging scheme~\cite{liu2017empower,ma2016end,lample2016neural,ratinov2009design,finkel2005incorporating}.
However, such design cannot deal with multi-label tokens.
Therefore, we customize the conventional CRF layer in LSTM-CRF~\cite{lample2016neural} into a Fuzzy CRF layer, which allows each token to have multiple labels without sacrificing computing efficiency.
% Besides that, we keep the structure of LSTM-CRF and refer the final model as Fuzzy CRF model.

% Going beyond the traditional
% sequence labeling 
% framework, we further propose a new prediction model, which is more robust to noisy labels introduced by distant supervision.
To adapt to imperfect labels generated by distant supervision, we go beyond the traditional sequence labeling framework and propose a new prediction model.
Specifically, instead of predicting the label of each single token, we propose to predict whether two adjacent tokens are tied in the same entity mention or not (i.e., broken).
The key motivation is that, even the boundaries of an entity mention are mismatched by distant supervision, most of its inner ties are not affected, and thus more robust to noise.
% Moreover, as the noisiest positive tokens, matched singleton entity mentions play the same role as non-entity tokens (i.e., negative labels) when we only care about ``Tie or Break''.
Therefore, we design a new \texttt{Tie or Break} tagging scheme to better exploit the noisy distant supervision.
Accordingly, we design a novel neural architecture that first forms all possible entity spans by detecting such ties, then identifies the entity type for each span.
The new scheme and neural architecture form our new model, \our, which proves to work better than the Fuzzy CRF model in our experiments.

% False positive is another issue in distant supervision labels.
% The full dictionary may contain many entities beyond the scope of the given corpus. 
% The aliases of such irrelevant entities may still be wrongly matched in the corpus, leading to false-positive labels.
% For example, when the dictionary has a non-related character name ``Wednesday Addams''\footnote{\url{https://en.wikipedia.org/wiki/Wednesday_Addams}} and its alias ``Wednesday'', many Wednesday's will be wrongly marked as persons.
% Therefore, we propose to conduct a corpus-aware dictionary tailoring to reduce false-positive labels.

We summarize our major contributions as
\begin{itemize}[noitemsep,leftmargin=*]
    \item We propose \our, a novel neural model with the new \texttt{Tie or Break} scheme for the distantly supervised NER task.
    \item We revise the traditional NER model to the Fuzzy-LSTM-CRF model, which serves as a strong distantly supervised baseline.
    \item We explore to refine distant supervision for better NER performance, such as incorporating high-quality phrases to reduce false-negative labels, 
    and conduct ablation experiments to verify the effectiveness.
    \item Experiments on three benchmark datasets demonstrate that \our achieves the best performance when only using dictionaries with no additional human effort and is even competitive with the supervised benchmarks.
\end{itemize}

We release all code and data for future studies\footnote{ \url{https://github.com/shangjingbo1226/AutoNER}}. Related open tools can serve as the NER module of various domain-specific systems in a plug-in-and-play manner.

%% file: 3-overview.tex
%!Tex Root = 0_main.tex
\section{Overview}

    Our goal, in this paper, is to learn a named entity tagger using, and only using dictionaries.
    Each dictionary entry consists of 1) the surface names of the entity, including a canonical name and a list of synonyms; and 2) the entity type. 
    Considering the limited coverage of dictionaries, we extend existing dictionaries by adding high-quality phrases as potential entities with unknown type.
    More details on refining distant supervision for better NER performance will be presented in Sec.~\ref{sec:dict_refine}.
    
    Given a raw corpus and a dictionary, we first generate entity labels (including unknown labels) by exact string matching, where conflicted matches are resolved by maximizing the total number of matched tokens~\cite{etzioni2005unsupervised,hanisch2005prominer,lin2012entity,he2017autoentity}.
    
    Based on the result of dictionary matching, each token falls into one of three categories: 
    1) it belongs to an entity mention with one or more known types; 
    2) it belongs to an entity mention with unknown type;
    and 3) It is marked as non-entity.
    
    Accordingly, we design and explore two neural models, Fuzzy-LSTM-CRF with the modified \texttt{IOBES} scheme and \our with the \texttt{Tie or Break} scheme, to learn named entity taggers based on such labels with unknown and multiple types.
    We will discuss the details in Sec.~\ref{sec:neural_models}.

%% file: 4-models.tex
%!Tex Root = 0_main.tex

\begin{figure*}[ht!]
    \centering
    \includegraphics[width=0.9\textwidth]{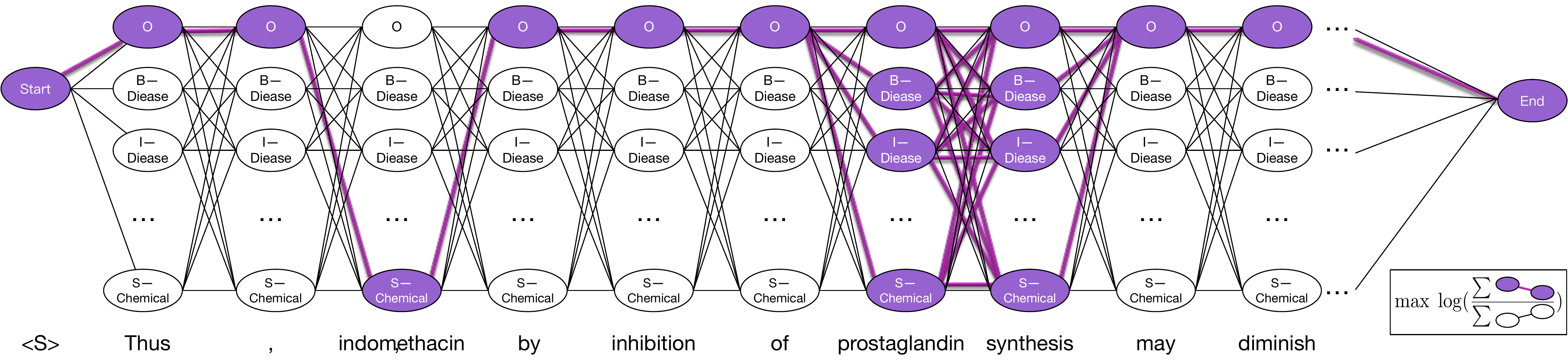}
    \vspace{-0.1cm}
    \caption{The illustration of the Fuzzy CRF layer with modified \texttt{IOBES} tagging scheme. The named entity types are \{\texttt{Chemical}, \texttt{Disease}\}. ``indomethacin'' is a matched \texttt{Chemical} entity and ``prostaglandin synthesis'' is an unknown-typed high-quality phrase.
    Paths from \texttt{Start} to \texttt{End} marked as purple form all possible label sequences given the distant supervision.}
    \label{fig:fcrf}
    \vspace{-0.3cm}
\end{figure*}

\section{Neural Models}
\label{sec:neural_models}
    In this section, we introduce two prediction models for the distantly supervised NER task, one under the traditional sequence labeling framework and another with a new labeling scheme.

    \subsection{Fuzzy-LSTM-CRF with Modified IOBES}
    \label{sec:fcrf}
        
        State-of-the-art named entity taggers follow the sequence labeling framework using \texttt{IOB} or \texttt{IOBES} scheme~\cite{ratinov2009design}, thus requiring a conditional random field (CRF) layer to capture the dependency between labels.
        However, both the original scheme and the conventional CRF layer cannot handle multi-typed or unknown-typed tokens.
        Therefore, we propose the modified \texttt{IOBES} scheme and Fuzzy CRF layer accordingly, as illustrated in Figure~\ref{fig:fcrf}.
        
        \smallsection{Modified IOBES}
        We define the labels according to the three token categories.
        1) For a token marked as one or more types, it is labeled with all these types and one of \{\texttt{I}, \texttt{B}, \texttt{E}, \texttt{S}\} according to its positions in the matched entity mention.
        2) For a token with unknown type, all five \{\texttt{I}, \texttt{O}, \texttt{B}, \texttt{E}, \texttt{S}\} tags are possible.
        Meanwhile, all available types are assigned.
        For example, when there are only two available types (e.g., Chemical and Disease), it has nine (i.e., $4 \times 2 + 1$) possible labels in total.
        3) For a token that is annotated as non-entity, it is labeled as \texttt{O}.
        
        As demonstrated in Fig.~\ref{fig:fcrf}, based on the dictionary matching results,
        ``indomethacin'' is a singleton \texttt{Chemical} entity and ``prostaglandin synthesis'' is an unknown-typed high-quality phrase.
        Therefore, ``indomethacin'' is labeled as \texttt{S-Chemical}, while both ``prostaglandin'' and ``synthesis'' are labeled as \texttt{O}, \texttt{B-Disease}, \texttt{I-Disease}, $\ldots$, and \texttt{S-Chemical} because the available entity types are \{\texttt{Chemical}, \texttt{Disease}\}.
        The non-entity tokens, such as ``Thus'' and ``by'', are labeled as \texttt{O}.
        
        \smallsection{Fuzzy-LSTM-CRF}
        We revise the LSTM-CRF model~\cite{lample2016neural} to the Fuzzy-LSTM-CRF model to support the modified \texttt{IOBES} labels.

        Given a word sequence $(X_1, X_2, \ldots, X_n)$, it is first passed through a word-level BiLSTM~\cite{hochreiter1997long} (i.e., forward and backward LSTMs).
        After concatenating the representations from both directions, the model makes independent tagging decisions for each output label.
        In this step, the model estimates the score $P_{i, y_j}$ for the word $X_i$ being the label $y_j$.
        
        We follow previous works~\cite{liu2017empower,ma2016end,lample2016neural} to define the score of the predicted sequence, the score of the predicted sequence $(y_1, y_2, \ldots, y_n)$ is defined as:
        \begin{equation} \label{eq:score}
            s(X, y) = \sum_{i = 0}^{n} \Phi_{y_i, y_{i + 1}} + \sum_{i = 1}^{n} P_{i, y_i}
        \end{equation}
        where, $\Phi_{y_i, y_{i + 1}}$ is the transition probability from a label $y_i$ to its next label $y_{i + 1}$.
        $\Phi$ is a $(k + 2) \times (k + 2)$ matrix, where $k$ is the number of distinct labels.
        Two additional labels \texttt{start} and \texttt{end} are used (only used in the CRF layer) to represent the beginning and end of a sequence, respectively. 
        
        The conventional CRF layer maximizes the probability of the only valid label sequence.
        However, in the modified \texttt{IOBES} scheme, one sentence may have multiple valid label sequences, as shown in Fig.~\ref{fig:fcrf}.
        Therefore, we extend the conventional CRF layer to a fuzzy CRF model.
        Instead, it maximizes the total probability of all possible label sequences by enumerating both the \texttt{\texttt{IOBES}} tags and all matched entity types.
        Mathematically, we define the optimization goal as Eq.~\ref{eq:fuzzy_crf_goal}.
        \begin{equation}
        \label{eq:fuzzy_crf_goal}
            p(y|X) = \frac{\sum_{\tilde{y} \in Y_{possible}} e^{s(X,\tilde{y})} }{\sum_{\tilde{y} \in Y_X} e^{s(X,\tilde{y})}}
        \end{equation}
        where $Y_X$ means all the possible label sequences for sequence $X$, and $Y_{possible}$ contains all the possible label sequences given the labels of modified \texttt{IOBES} scheme.
        Note that, when all labels and types are known and unique, the fuzzy CRF model is equivalent to the conventional CRF.
        
        During the training process, we maximize the log-likelihood function of Eq.~\ref{eq:fuzzy_crf_goal}. 
        For inference, we apply the Viterbi algorithm to maximize the score of Eq.~\ref{eq:score} for each input sequence. 
        
\begin{figure*}[ht!]
    \centering
    \includegraphics[width=0.95\textwidth]{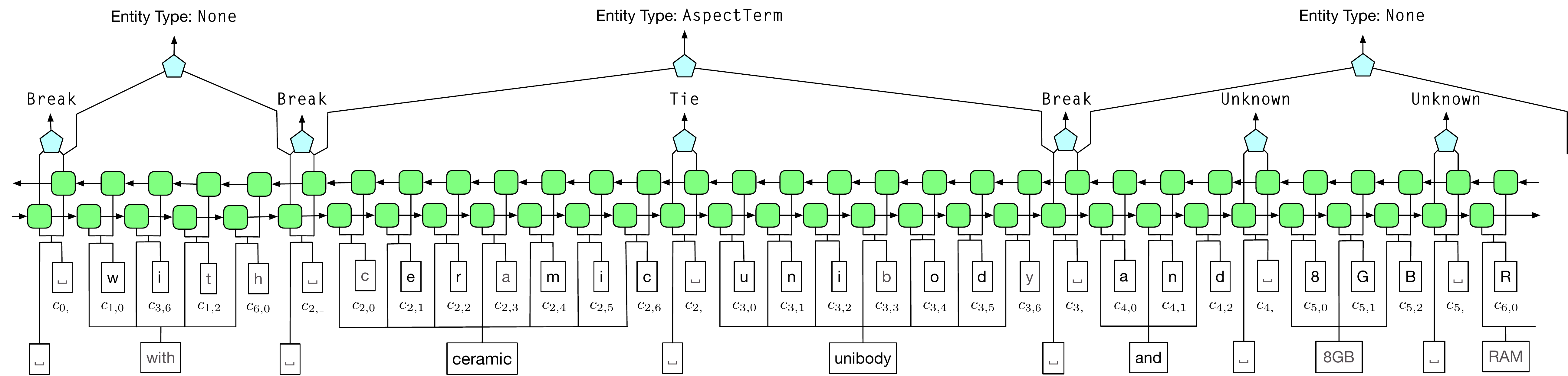}
    \vspace{-0.1cm}
    \caption{The illustration of \our with \texttt{Tie or Break} tagging scheme. 
    The named entity type is \{\texttt{AspectTerm}\}. 
    ``ceramic unibody'' is a matched \texttt{AspectTerm} entity and ``8GB RAM'' is an unknown-typed high-quality phrase. \texttt{Unknown} labels will be skipped during the model training.}\label{fig:autoner}
    \vspace{-0.3cm}
\end{figure*}

    \subsection{\our with ``Tie or Break''}
        Identifying the nature of the distant supervision, we go beyond the sequence labeling framework and propose a new tagging scheme, \texttt{Tie or Break}.
        It focuses on the ties between adjacent tokens, i.e., whether they are tied in the same entity mentions or broken into two parts.
        Accordingly, we design a novel neural model for this scheme.

        \smallsection{``Tie or Break'' Tagging Scheme}
        Specifically, for every two adjacent tokens, the connection between them is labeled as 
        (1) \texttt{Tie}, when the two tokens are matched to the same entity; 
        (2) \texttt{Unknown}, if at least one of the tokens belongs to an unknown-typed high-quality phrase;
        (3) \texttt{Break}, otherwise.
        
        An example can be found in Fig.~\ref{fig:autoner}.
        The distant supervision shows that ``ceramic unibody'' is a matched \texttt{AspectTerm} and ``8GB RAM'' is an unknown-typed high-quality phrase.
        Therefore, a \texttt{Tie} is labeled between ``ceramic'' and ``unibody'', while \texttt{Unknown} labels are put before ``8GB'', between ``8GB'' and ``RAM'', and after ``RAM''.

        Tokens between every two consecutive \texttt{Break} form a token span.
        Each token span is associated with all its matched types, the same as for the modified \texttt{IOBES} scheme.
        For those token spans without any associated types, such as ``with'' in the example, we assign them the additional type \texttt{None}.
        
        We believe this new scheme can better exploit the knowledge from dictionary according to the following two observations.
        First, even though the boundaries of an entity mention are mismatched by distant supervision, most of its inner ties are not affected.
        More interestingly, compared to multi-word entity mentions, matched unigram entity mentions are more likely to be false-positive labels. 
        However, such false-positive labels will not introduce incorrect labels with the \texttt{Tie or Break} scheme, since either the unigram is a true entity mention or a false positive, it always brings two \texttt{Break} labels around.
    
        \smallsection{\our}
        In the \texttt{Tie or Break} scheme, entity spans and entity types are encoded into two folds.
        Therefore, we separate the entity span detection and entity type prediction into two steps.
        
        For entity span detection, we build a binary classifier to distinguish \texttt{Break} from \texttt{Tie}, while \texttt{Unknown} positions will be skipped.
        Specifically, as shown in Fig.~\ref{fig:autoner}, for the prediction between $i$-th token and its previous token, we concatenate the output of the BiLSTM as a new feature vector, $\mathbf{u}_i$.
        $\mathbf{u}_i$ is then fed into a sigmoid layer, which estimates the probability that there is a \texttt{Break} as
        $$
            p(y_{i} = \mbox{\texttt{Break}} | \mathbf{u}_{i}) =  \sigma(\mathbf{w}^T \mathbf{u}_{i})
        $$
        where $y_{i}$ is the label between the $i$-th and its previous tokens, $\sigma$ is the sigmoid function, and $\mathbf{w}$ is the sigmoid layer's parameter.
        The entity span detection loss is then computed as follows.
        $$
            \mathcal{L}_{\mbox{span}} = \sum_{i | y_{i} \neq \mbox{\texttt{Unknown}}} l\big(y_{i}, p(y_i = \mbox{\texttt{Break}} | \mathbf{u}_i)\big)
        $$
        Here, $l(\cdot, \cdot)$ is the logistic loss.
        % Note that those \texttt{Unknown} positions are not considered in the loss function.
        Note that those \texttt{Unknown} positions are skipped.
        
        After obtaining candidate entity spans,
        we further identify their entity types, including the \texttt{None} type for non-entity spans.
        As shown in Fig.~\ref{fig:autoner}, the output of the BiLSTM will be re-aligned to form a new feature vector, which is referred as $\mathbf{v}_i$ for $i$-th span candidate.
        $\mathbf{v}_i$ will be further fed into a softmax layer, which estimates the entity type distribution as
        $$
            p(t_j | \mathbf{v}_i) = \frac{e^{\mathbf{t}_j^T\mathbf{v}_i}}{\sum_{t_k \in L} e^{\mathbf{t}_k^T\mathbf{v}_i}}
        $$
        where $t_j$ is an entity type and $L$ is the set of all entity types including \texttt{None}.
      
        Since one span can be labeled as multiple types, we mark the possible set of types for $i$-th entity span candidate as $L_i$.
        Accordingly, we modify the cross-entropy loss as follows.
        $$
            \mathcal{L}_{\mbox{type}} = H(\hat{p}(\cdot |\mathbf{v}_i, L_i), p(\cdot |\mathbf{v}_i))
        $$
        Here, $H(p, q)$ is the cross entropy between $p$ and $q$, and  $\hat{p}(t_j|\mathbf{v}_i, L_i)$ is the soft supervision distribution. Specifically, it is defined as:
        $$
            \hat{p}(t_j|\mathbf{v}_i, L_i) = \frac{\delta(t_j \in L_i) \cdot e^{\mathbf{t}_j^T\mathbf{v}_i}}{\sum_{t_k \in L} \delta(t_k \in L_i) \cdot e^{\mathbf{t}_k^T\mathbf{v}_i}}
        $$ 
        where $\delta(t_j \in L_i)$ is the boolean function of checking whether the $i$-th span candidate is labeled as the type $t_j$ in the distant supervision. 
    
        It's worth mentioning that \our has no CRF layer and Viterbi decoding, thus being more efficient than Fuzzy-LSTM-CRF for inference.

    \subsection{Remarks on ``Unknown'' Entities}
        ``Unknown'' entity mentions are not the entities of other types, but the tokens that we are less confident about their boundaries and/or cannot identify their types based on the distant supervision. 
        For example, in Figure 1, ``prostaglandin synthesis'' is an ``unknown'' token span. The distant supervision cannot decide whether it is a \texttt{Chemical}, a \texttt{Disease}, an entity of other types, two separate single-token entities, or (partially) not an entity. 
        Therefore, in the FuzzyCRF model, we assign all possible labels for these tokens.

        In our AutoNER model, these ``unknown'' positions have undefined boundary and type losses, because (1) they make the boundary labels unclear; and (2) they have no type labels. Therefore, they are skipped.

%% file: 5-dict.tex
%!Tex Root = 0_main.tex
\section{Distant Supervision Refinement}
\label{sec:dict_refine}
    In this section, we present two techniques to refine the distant supervision for better named entity taggers.
    Ablation experiments in Sec.~\ref{sec:each_component} verify their effectiveness empirically.
    
    \subsection{Corpus-Aware Dictionary Tailoring}
    
    In dictionary matching, blindly using the full dictionary may introduce false-positive labels, as there exist many entities beyond the scope of the given corpus but their aliases can be matched.
    For example, when the dictionary has a non-related character name ``Wednesday Addams''\footnote{\url{https://en.wikipedia.org/wiki/Wednesday_Addams}} and its alias ``Wednesday'', many Wednesday's will be wrongly marked as persons.
    % A larger dictionary may lead to a relatively higher recall, but usually hurts the precision severely.
    In an ideal case, the dictionary should cover, and only cover entities occurring in the given corpus to ensure a high precision while retaining a reasonable coverage.
    
    As an approximation, we tailor the original dictionary to a corpus-related subset by excluding entities whose canonical names never appear in the given corpus.
    The intuition behind is that to avoid ambiguities, people will likely mention the canonical name of the entity at least once.
    For example, in the biomedical domain, this is true for $88.12\%$, $95.07\%$ of entity mentions on the BC5CDR and NCBI datasets respectively.
    We expect the NER model trained on such tailored dictionary will have a higher precision and a reasonable recall compared to that trained on the original dictionary.

    \subsection{Unknown-Typed High-Quality Phrases}
    
    Another issue of the distant supervision is about the false-negative labels.
    When a token span cannot be matched to any entity surface names in the dictionary, because of the limited coverage of dictionaries, it is still difficult to claim it as non-entity (i.e., negative labels) for sure.
    Specifically, some high-quality phrases out of the dictionary may also be potential entities.
    
    % To address this issue, we apply a distantly supervised phrase mining method~\cite{shang2017automated} to extract high-quality, domain-specific phrases from the corpus.
    We utilize the state-of-the-art distantly supervised phrase mining method, AutoPhrase~\cite{shang2017automated}, with the corpus and dictionary in the given domain as input.
    AutoPhrase only requires unlabeled text and a dictionary of high-quality phrases.
    We obtain quality multi-word and single-word phrases by posing thresholds (e.g., 0.5 and 0.9 respectively). 
    In practice, one can find more unlabeled texts from the same domain (e.g., PubMed papers and Amazon laptop reviews) and use the same domain-specific dictionary for the NER task.
    In our experiments, for the biomedical domain, we use the titles and abstracts of  686,568 PubMed papers (about $4\%$) uniformly sampled from the whole PubTator database as the training corpus.
    For the laptop review domain, we use the Amazon laptop review dataset\footnote{\url{http://times.cs.uiuc.edu/~wang296/Data/}}, which is designed for the aspect-based sentiment analysis ~\cite{wang2011latent}.
    
    We treat out-of-dictionary phrases as potential entities with ``unknown'' type and incorporate them as new dictionary entries.
    After this, only token spans that cannot be matched in this extended dictionary will be labeled as non-entity.
    Being aware of these high-quality phrases, we expect the trained NER tagger should be more accurate.

%% file: 6-exp.tex
%!Tex Root = 0_main.tex

\section{Experiments}

    We conduct experiments on three benchmark datasets to evaluate and compare our proposed Fuzzy-LSTM-CRF and \our with many other methods.
    We further investigate the effectiveness of our proposed refinements for the distant supervision and the impact of the number of distantly supervised sentences.
    
\begin{table}[t]
\centering
    \caption{Dataset Overview.}
    \label{tbl:dataset}
\scalebox{0.6}{
    \begin{tabularx}{0.77\textwidth}{ccccc}
    \toprule
    Dataset & BC5CDR & NCBI-Disease & LaptopReview \\
    \midrule
    \midrule
    Domain & Biomedical & Biomedical & Technical Review\\
    \midrule
    Entity Types & \texttt{Disease}, \texttt{Chemical} & \texttt{Disease} & \texttt{AspectTerm} \\
    \midrule
    Dictionary & MeSH + CTD & MeSH + CTD & Computer Terms\\
    \midrule
    Raw Sent. \# & 20,217 & 7,286 & 3,845 \\
    \bottomrule
    \end{tabularx}
}
    \vspace{-0.4cm}
\end{table}

\begin{table*}[t]
\centering
    \caption{[Biomedical Domain] NER Performance Comparison. The supervised benchmarks on the BC5CDR and NCBI-Disease datasets are LM-LSTM-CRF and LSTM-CRF respectively ~\cite{wang2018cross}. SwellShark has no annotated data, but for entity span extraction, it requires pre-trained POS taggers and extra human efforts of designing POS tag-based regular expressions and/or hand-tuning for special cases.}
    \label{tbl:ner}
    \scalebox{1}{
        \small
        \begin{tabularx}{0.97\textwidth}{cccccccc}
        \toprule
        \multirow{2}{*}{Method} & Human Effort & \multicolumn{3}{c}{BC5CDR}  & \multicolumn{3}{c}{NCBI-Disease} \\
        \cmidrule{3-8}
         & other than Dictionary & Pre & Rec & F1 & Pre & Rec & F1 \\
        \midrule
        \midrule
        Supervised Benchmark & Gold Annotations & 88.84 & 85.16 & \textbf{86.96} & 86.11 & 85.49 & \textbf{85.80} \\
        % LM-LSTM-CRF & \multirow{4}{*}{Gold Annotations} & 88.84 & 85.16 & \textbf{86.96} & 84.95 & 82.92 & 83.92 \\
        % \cmidrule{3-8}
        % LSTM-CRF & & 87.60 & 86.25 & 86.92 & 86.11 & 85.49 & \textbf{85.80} \\
        % \cmidrule{3-8}
        % LSTM-CNN-CRF & & 89.16 & 84.28 & 86.65 & 86.89 & 78.75 & 82.62 \\
        % \cmidrule{3-8}
        % TaggerOne & & 88.23 & 80.97 & 84.44 & 83.5 & 79.6 & 81.5 \\
        \midrule
        \midrule
        \multirow{2}{*}{SwellShark} & Regex Design + Special Case Tuning  & 86.11 & 82.39 & 84.21 & 81.6 & 80.1 & \textbf{80.8} \\
        \cmidrule{2-8}
         & Regex Design  & 84.98 & 83.49 & \textbf{84.23} & 64.7 & 69.7 & 67.1 \\
        \midrule
        \midrule
        Dictionary Match & \multirow{3}{*}{None} & 93.93 & 58.35 & 71.98 & 90.59 & 56.15 & 69.32\\
        \cmidrule{3-8}
        Fuzzy-LSTM-CRF &  & 88.27 & 76.75 & 82.11 & 79.85 & 67.71 & 73.28 \\
        \cmidrule{3-8}
        \our &  & 88.96 & 81.00 & \textbf{84.8} & 79.42 & 71.98 & \textbf{75.52} \\
        \bottomrule
        \end{tabularx}
    }
    \vspace{-0.4cm}
\end{table*}

    \subsection{Experimental Settings}
    
        \noindent\textbf{Datasets} are briefly summarized in Table~\ref{tbl:dataset}. More details as as follows.
        \begin{itemize}[noitemsep,leftmargin=*]
            \item \textbf{BC5CDR} is from the most recent BioCreative V Chemical and Disease Mention Recognition task. It has 1,500 articles containing 15,935 \texttt{Chemical} and 12,852 \texttt{Disease} mentions. 
            \item \textbf{NCBI-Disease} focuses on Disease Name Recognition. It contains 793 abstracts and 6,881 \texttt{Disease} mentions.
            \item \textbf{LaptopReview} is from the SemEval 2014 Challenge, Task 4 Subtask 1~\cite{semeval2014task4} focusing on laptop aspect term (e.g., ``disk drive'') Recognition.
            It consists of 3,845 review sentences and 3,012 \texttt{AspectTerm} mentions.
            % We use the laptop dataset because we can find a domain-specific dictionary.
        \end{itemize}
        All datasets are publicly available.
        The first two datasets are already partitioned into three subsets: a training set, a development set, and a testing set. 
        For the LaptopReview dataset, we follow~\cite{giannakopoulos2017unsupervised} and randomly select 20\% from the training set as the development set.
        Only raw texts are provided as the input of distantly supervised models, while the gold training set is used for supervised models.
        
        \smallsection{Domain-Specific Dictionary}
        For the biomedical datasets, the dictionary is a combination of both the MeSH database\footnote{\url{https://www.nlm.nih.gov/mesh/download_mesh.html}} and the CTD Chemical and Disease vocabularies\footnote{\url{http://ctdbase.org/downloads/}}. 
        The dictionary contains 322,882 \texttt{Chemical} and \texttt{Disease} entity surfaces.
        % After tailored, there are 2,482 and 931 typed surfaces left for BC5CDR and NCBI-disease datasets respectively.
        For the laptop review dataset, the dictionary has 13,457 computer terms crawled from a public website\footnote{\url{https://www.computerhope.com/jargon.htm}}.
        % The tailored dictionary contains 272 \texttt{AspectTerm}'s.
        
        \smallsection{Metric}
        We use the micro-averaged $F_1$ score as the evaluation metric.
        Meanwhile, precision and recall are presented. 
        The reported scores are the mean across five different runs. 
            
        \smallsection{Parameters and Model Training}
        Based on the analysis conducted in the development set, we conduct optimization with the stochastic gradient descent with momentum.
        We set the batch size and the momentum to $10$ and $0.9$. 
        The learning rate is initially set to $0.05$ and will be shrunk by $40\%$ if there is no better development $F_1$ in the recent $5$ rounds.
        Dropout of a ratio $0.5$ is applied in our model.
        For a better stability, we use gradient clipping of $5.0$.
        Furthermore, we employ the early stopping in the development set. 
        
        \smallsection{Pre-trained Word Embeddings}
        For the biomedical datasets, we use the pre-trained 200-dimension word vectors \footnote{\url{http://bio.nlplab.org/}}
        from~\cite{moen2013distributional}, which are trained on the whole PubMed abstracts, all the full-text articles from PubMed Central (PMC), and English Wikipedia. For the laptop review dataset, we use the GloVe 100-dimension pre-trained word vectors\footnote{\url{https://nlp.stanford.edu/projects/glove/}} instead, which are trained on the Wikipedia and GigaWord.
  
    \subsection{Compared Methods}
        \noindent\textbf{Dictionary Match} is our proposed distant supervision generation method.
        Specifically, we apply it to the testing set directly to obtain entity mentions with exactly the same surface name as in the dictionary. 
        The type is assigned through a majority voting.
        By comparing with it, we can check the improvements of neural models over the distant supervision itself.
        
        % \noindent\textbf{Fuzzy-LSTM-CRF} with the modified \texttt{IOBES} scheme, as described in Sec.~\ref{sec:fcrf}, serves as a strong distantly supervised baseline.
        
        \smallskip
        \noindent\textbf{SwellShark}, in the biomedical domain, is arguably the best distantly supervised model, especially on the BC5CDR and NCBI-Disease datasets~\cite{fries2017swellshark}.
        It needs no human annotated data, however, it requires extra expert effort for entity span detection on building POS tagger, designing effective regular expressions, and hand-tuning for special cases.
        
        \smallskip
        \noindent\textbf{Distant-LSTM-CRF} achieved the best performance on the LaptopReview dataset without annotated training data using a distantly supervised LSTM-CRF model~\cite{giannakopoulos2017unsupervised}.
        
        \smallskip
        \noindent\textbf{Supervised benchmarks} on each dataset are listed to check whether \our can deliver competitive performance. On the BC5CDR and NCBI-Disease datasets, LM-LSTM-CRF~\cite{liu2017empower} and LSTM-CRF~\cite{lample2016neural} achieve the state-of-the-art $F_1$ scores without external resources, respectively~\cite{wang2018cross}.
        On the LaptopReview dataset, we present the scores of the Winner in the SemEval2014 Challenge Task 4 Subtask 1~\cite{semeval2014task4}.

\begin{table}[t]
\centering
    \caption{[Technical Review Domain] NER Performance Comparison. The supervised benchmark refers to the challenge winner.}
    \label{tbl:ner_laptop}
    \scalebox{1}{
        \small
        \begin{tabularx}{0.9\columnwidth}{cccc}
        \toprule
        \multirow{2}{*}{Method} & \multicolumn{3}{c}{LaptopReview} \\
        \cmidrule{2-4}
        & Pre & Rec & F1 \\
        \midrule
        \midrule
        Supervised Benchmark & 84.80 & 66.51 & \textbf{74.55} \\
        % \midrule
        % Supervised LSTM-CRF & - & - & \textbf{77.96} \\
        \midrule
        \midrule
        Distant-LSTM-CRF & 74.03 & 31.59 & 53.93\\
        \midrule
        Dictionary Match & 90.68 & 44.65 & 59.84 \\
        \midrule
        Fuzzy-LSTM-CRF & 85.08 & 47.09 & 60.63 \\
        \midrule
        \our & 72.27 & 59.79 & \textbf{65.44} \\
        \bottomrule
        \end{tabularx}
    }
    \vspace{-0.4cm}
\end{table}

    \subsection{NER Performance Comparison}
        We present $F_1$, precision, and recall scores on all datasets in Table~\ref{tbl:ner} and Table~\ref{tbl:ner_laptop}.
        From both tables, one can find the \our achieves the best performance when there is no extra human effort.
        Fuzzy-LSTM-CRF does have some improvements over the Dictionary Match, but it is always worse than \our.
        
        Even though SwellShark is designed for the biomedical domain and utilizes much more expert effort, \our outperforms it in almost all cases.
        The only outlier happens on the NCBI-disease dataset when the entity span matcher in SwellShark is carefully tuned by experts for many special cases.
        
        It is worth mentioning that \our beats Distant-LSTM-CRF, which is the previous state-of-the-art distantly supervised model on the LaptopReview dataset.
        
        Moreover, \our's performance is competitive to the supervised benchmarks. 
        % For example, on the BC5CDR dataset, its $F_1$ score is only $2.58\%$ away from the supervised benchmark.
        For example, on the BC5CDR dataset, its $F_1$ score is only $2.16\%$ away from the supervised benchmark.

        % The $F_1$ score of \our on the LaptopReview dataset is relatively low because the aspect terms are not that well-defined, compared to the Disease and Chemical. 

\begin{table*}[ht!]
\centering
    \caption{Ablation Experiments for Dictionary Refinement. The dictionary for the LaptopReview dataset contains no alias, so the corpus-aware dictionary tailoring is not applicable.}
    \label{tbl:each_component}
    \scalebox{1}{
        \small
        \begin{tabularx}{0.96\textwidth}{cccccccccc}
        \toprule
        \multirow{2}{*}{Method} & \multicolumn{3}{c}{BC5CDR}  & \multicolumn{3}{c}{NCBI-Disease} & \multicolumn{3}{c}{LaptopReview}\\
        \cmidrule{2-10}
          & Pre & Rec & F1 & Pre & Rec & F1 & Pre & Rec & F1 \\
        \midrule
        \midrule
        \our w/ Original Dict & 82.79 & 70.40 & 76.09 & 53.14 & 63.54 & 57.87 & 69.96 & 49.85 & 58.21\\
        \midrule
        \our w/ Tailored Dict & 84.57 & 70.22 & 76.73 & 77.30 & 58.54 & 66.63 & \multicolumn{3}{c}{Not Applicable}  \\
        \midrule
        \our w/ Tailored Dict \& Phrases & 88.96 & 81.00 & \textbf{84.8} & 79.42 & 71.98 & \textbf{75.52} & 72.27 & 59.79 & \textbf{65.44} \\
        \bottomrule
        \end{tabularx}
    }
    \vspace{-0.4cm}
\end{table*}

\begin{figure*}[ht!]
    \centering
    \subfigure[BC5CDR]{ \includegraphics[width=0.3\textwidth]{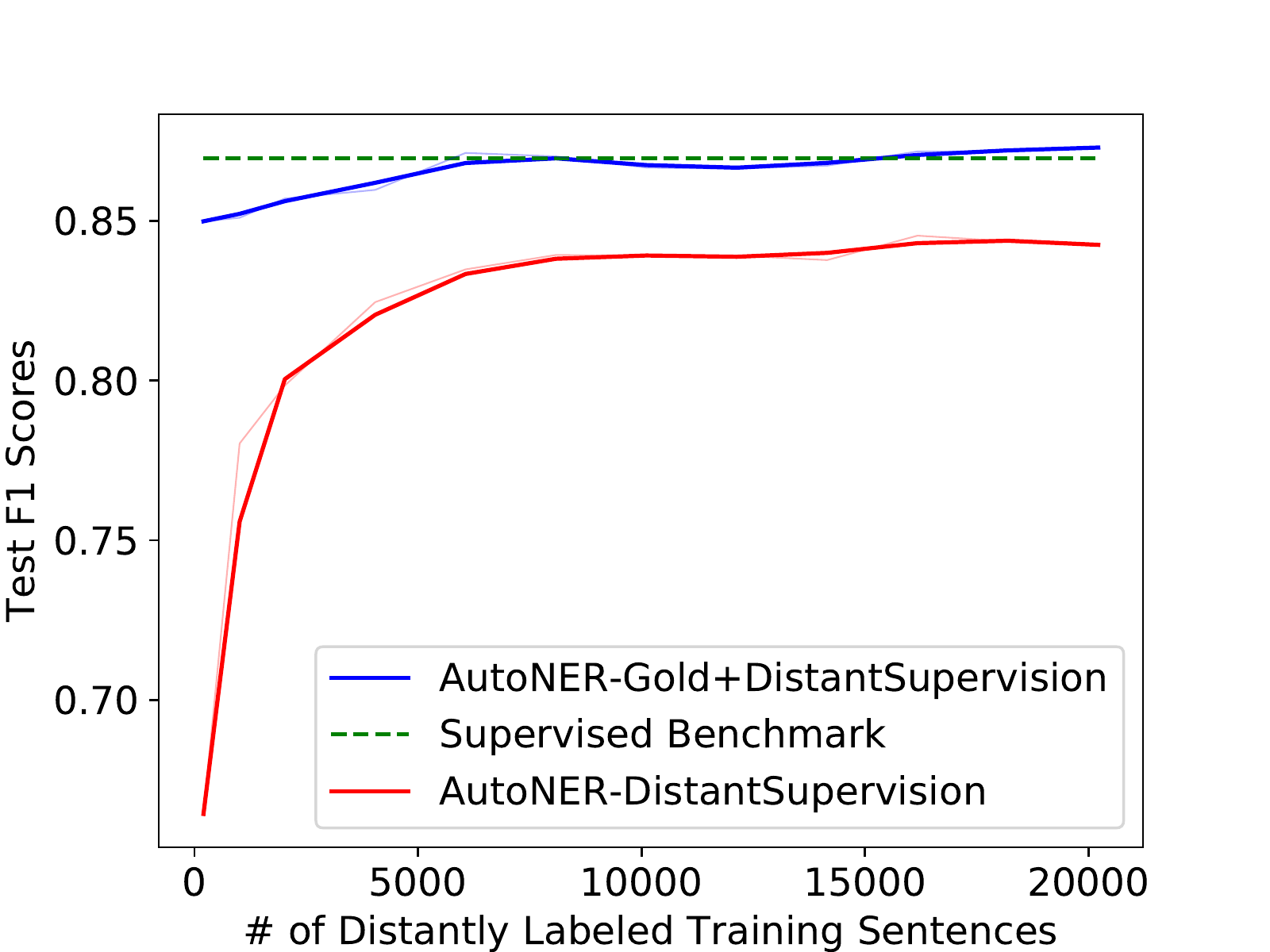}
    }
    \subfigure[NCBI]{ \includegraphics[width=0.3\textwidth]{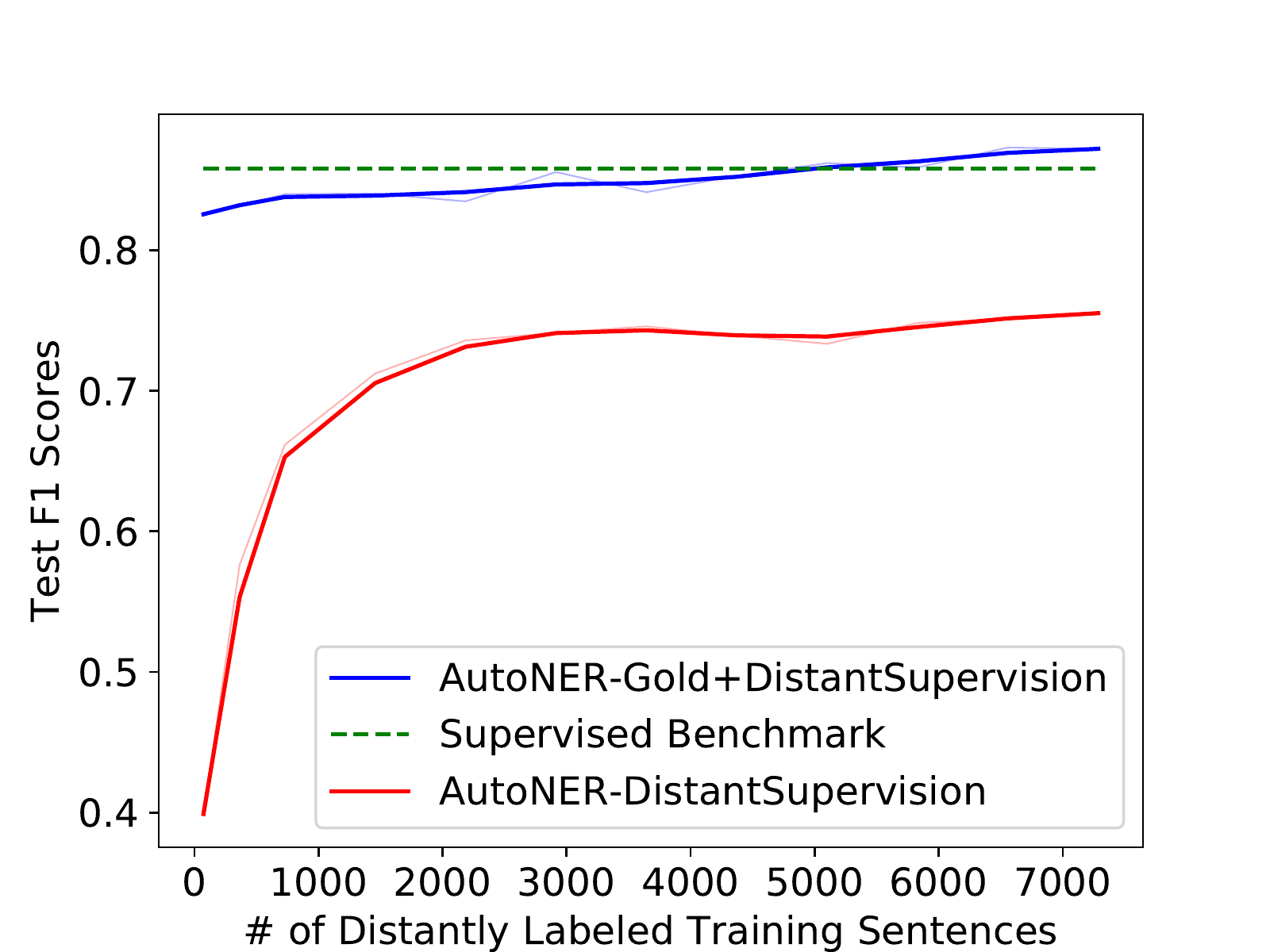}
    }
    \subfigure[LaptopReview]{ \includegraphics[width=0.3\textwidth]{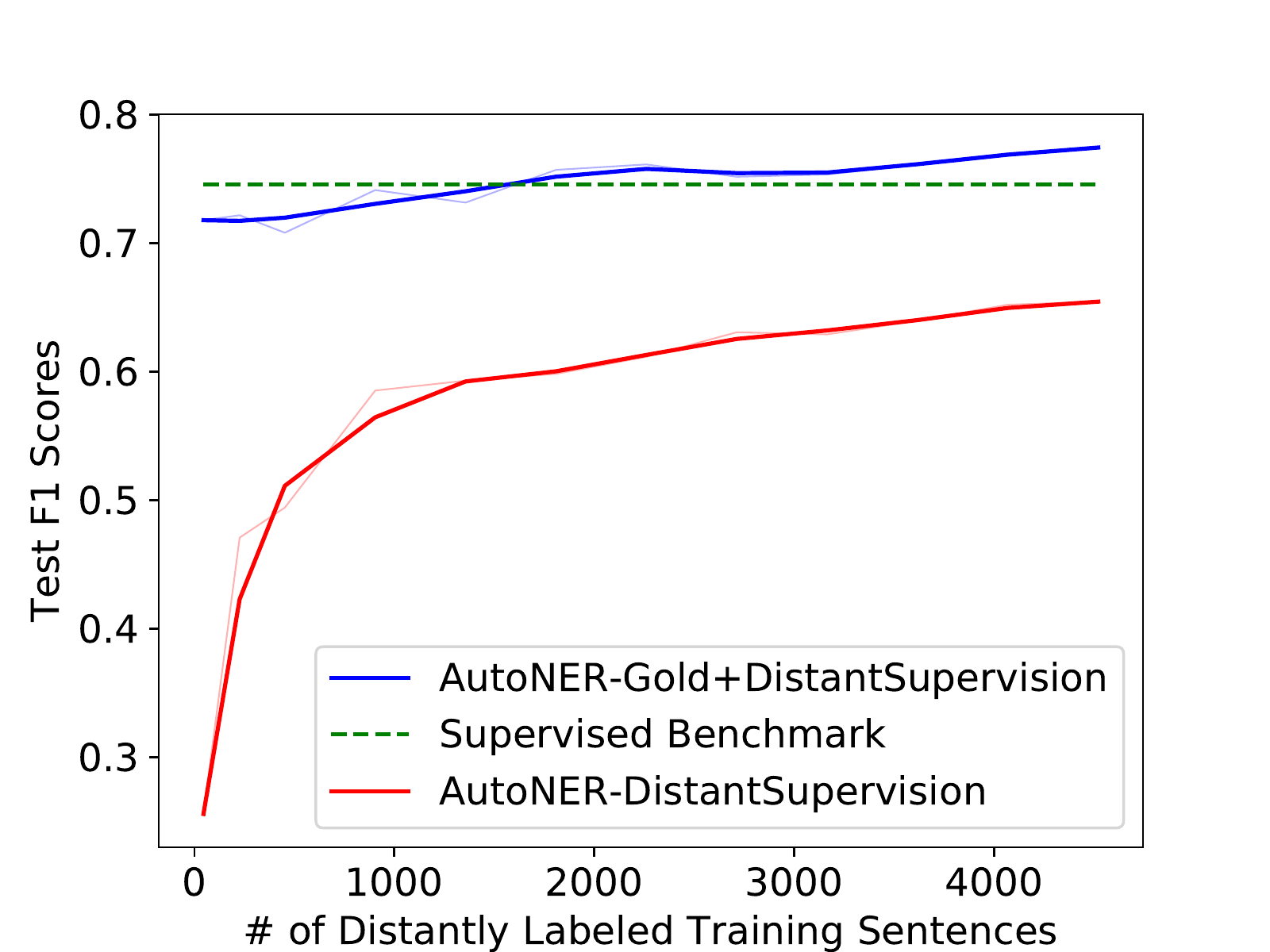}
    }
    \vspace{-0.3cm}
    \caption{\our: Test $F_1$ score vs. the number of distantly supervised sentences.}\label{fig:f1_vs_data}
    \vspace{-0.4cm}
\end{figure*}

    \subsection{Distant Supervision Explorations}
        \label{sec:each_component}
        
        % We explore how to refine distant supervision for a more effective \our model.
        We investigate the effectiveness of the two techniques that we proposed in Sec.~\ref{sec:dict_refine} via ablation experiments.
        As shown in Table~\ref{tbl:each_component}, 
        using the tailored dictionary always achieves better $F_1$ scores than using the original dictionary.
        By using the tailored dictionary, the precision of the \our model will be higher, while the recall will be retained similarly. 
        For example, on the NCBI-Disease dataset, it significantly boosts the precision from $53.14\%$ to $77.30\%$ with an acceptable recall loss from $63.54\%$ to $58.54\%$.
        Moreover, incorporating unknown-typed high-quality phrases in the dictionary enhances every score of \our models significantly, especially the recall.
        These results match our expectations well.

    \subsection{Test $F_1$ Scores vs. Size of Raw Corpus}

        Furthermore, we explore the change of test $F_1$ scores when we have different sizes of distantly supervised texts.
        We sample sentences uniformly random from the given raw corpus and then evaluate \our models trained on the selected sentences.
        We also study what will happen when the gold training set is available.
        The curves can be found in Figure~\ref{fig:f1_vs_data}.
        The X-axis is the number of distantly supervised training sentences while the Y-axis is the $F_1$ score on the testing set.
        
        When using distant supervision only, one can observe a significant growing trend of test $F_1$ score in the beginning, but later the increasing rate slows down when there are more and more raw texts.
        
        When the gold training set is available, the distant supervision is still helpful to \our.
        In the beginning, \our works worse than the supervised benchmarks.
        Later, with enough distantly supervised sentences, \our outperforms the supervised benchmarks.
        We think there are two possible reasons: (1) The distant supervision puts emphasis on those matchable entity mentions; and (2) The gold annotation may miss some good but matchable entity mentions. 
        These may guide the training of \our to a more generalized model, and thus have a higher test $F_1$ score.

    \subsection{Comparison with Gold Supervision}
        To demonstrate the effectiveness of distant supervision, we try to compare our method with gold annotations provided by human experts.
        
\begin{figure}[t!]
    \centering
    \includegraphics[width=0.33\textwidth]{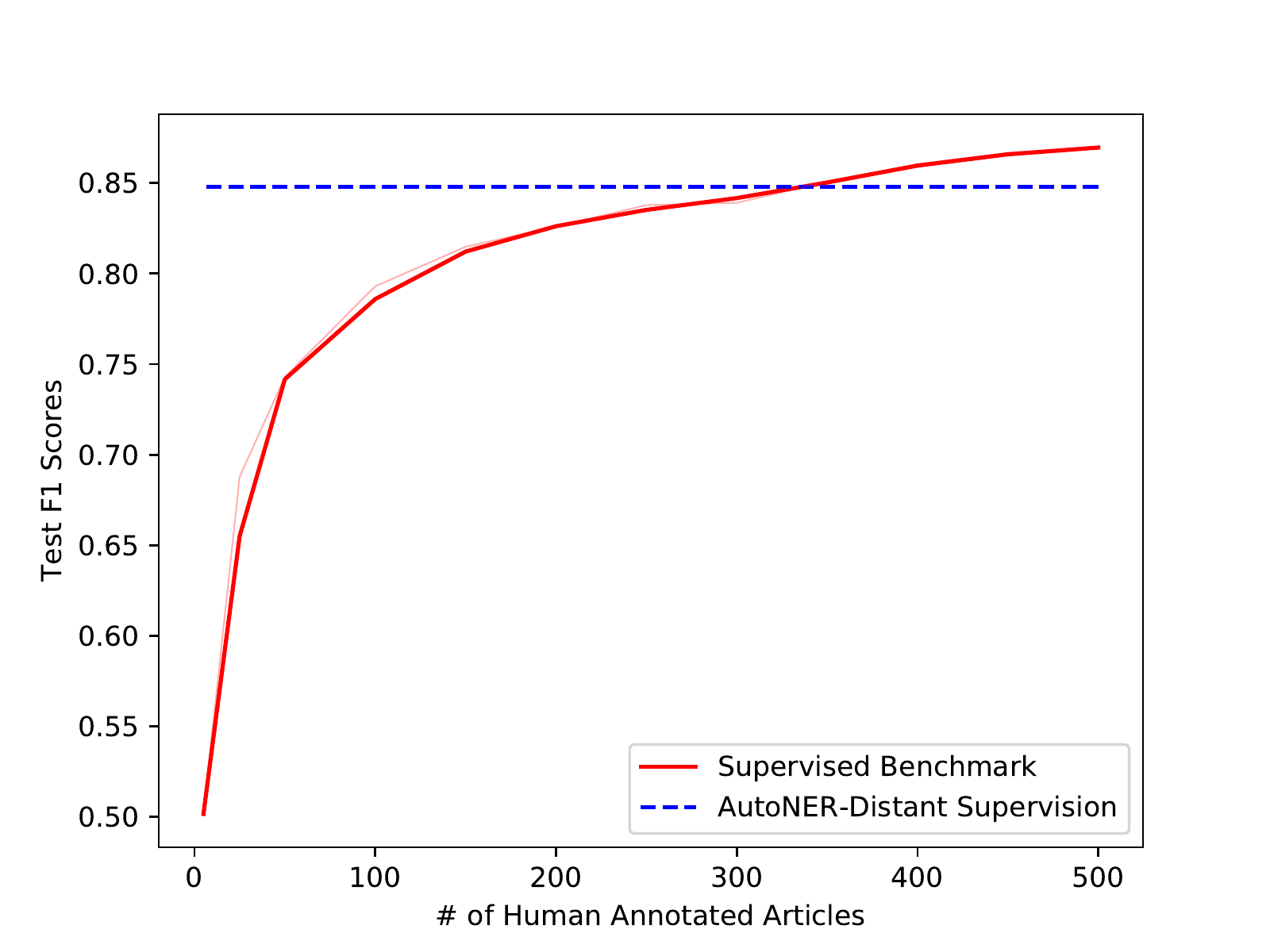}
    \vspace{-0.3cm}
    \caption{\our: Test $F_1$ score vs. the number of human annotated articles.}\label{fig:f1_vs_supervision}
    \vspace{-0.3cm}
\end{figure}
        
        Specifically, we conduct experiments on the BC5CDR dataset by sampling different amounts of annotated articles for model training.
        As shown in Figure~\ref{fig:f1_vs_supervision}, we found that our method outperforms the supervised method by a large margin when less training examples are available.
        For example, when there are only 50 annotated articles available, the test F1 score drops substantially to 74.29\%. 
        To achieve a similar test F1 score (e.g., 83.91\%) as our AutoNER model’s (i.e., 84.8\%), the supervised benchmark model requires at least 300 annotated articles. 
        Such results indicate the effectiveness and usefulness of AutoNER on the scenario without sufficient human annotations.
        
        Still, we observe that, when the supervised benchmark is trained with all annotations, it achieves the performance better than AutoNER.
        We conjugate that this is because AutoNER lacks more advanced techniques to handle distant supervision, and we leave further improvements of AutoNER to the future work.

%% file: 2-related.tex
\section{Related Work}

    The task of supervised named entity recognition (NER) is typically embodied as a sequence labeling problem.
    Conditional random fields (CRF) models built upon human annotations and handcrafted features are the standard~\cite{finkel2005incorporating,settles2004biomedical,leaman2008banner}.
    Recent advances in neural models have freed domain experts from handcrafting features for NER tasks.
    ~\cite{lample2016neural,ma2016end,liu2017empower}.
    Such neural models are increasingly common in the domain-specific NER tasks~\cite{sahu2016recurrent,dernoncourt2017identification,wang2018cross}.
    Semi-supervised methods have been explored to further improve the accuracy by either augmenting labeled datasets with word embeddings or bootstrapping techniques in tasks like gene name recognition~\cite{kuksa2010semi,tang2014evaluating,vlachos2006bootstrapping}. 
    Unlike these existing approaches, our study focuses on the distantly supervised setting without any expert-curated training data.
    
    Distant supervision has attracted many attentions to alleviate human efforts.
    Originally, it was proposed to leverage knowledge bases to supervise relation extraction tasks~\cite{craven1999constructing,mintz2009distant}. 
    AutoPhrase has demonstrated powers in extracting high-quality phrases from domain-specific corpora like scientific papers and business reviews~\cite{shang2017automated} but it cannot categorize phrases into typed entities in a context-aware manner. 
    We incorporate the high-quality phrases to enrich the domain-specific dictionary.
    
    There are attempts on the distantly supervised NER task recently~\cite{ren2015clustype,fries2017swellshark,he2017autoentity,giannakopoulos2017unsupervised}. 
    For example, SwellShark~\cite{fries2017swellshark}, specifically designed for biomedical NER, leverages a generative model to unify and model noise across different supervision sources for named entity typing.
    However, it leaves the named entity span detection to a heuristic combination of dictionary matching and part-of-speech tag-based regular expressions, which require extensive expert effort to cover many special cases.
    Other methods~\cite{ren2015clustype,he2017autoentity} also utilize similar approaches to extract entity span candidates before entity typing.
    Distant-LSTM-CRF~\cite{giannakopoulos2017unsupervised} has been proposed for the distantly supervised aspect term extraction, which can be viewed as an entity recognition task of a single type for business reviews. 
    As shown in our experiments, our models can outperform Distant-LSTM-CRF significantly on the laptop review dataset. 

    To the best of our knowledge, \our is the most effective model that can learn NER models by using, and only using dictionaries without any additional human effort.

%% file: 7-con.tex
%!Tex Root = 0_main.tex
\section{Conclusion and Future Work}

In this paper, we explore how to learn an effective NER model by using, and only using dictionaries.
We design two neural architectures, Fuzzy-LSTM-CRF model with a modified \texttt{IOBES} tagging scheme and \our with a new \texttt{Tie or Break} scheme.
In experiments on three benchmark datasets, \our achieves the best $F_1$ scores without additional human efforts. 
Its performance is even competitive to the supervised benchmarks with full human annotation.
In addition, we discuss how to refine the distant supervision for better NER performance, including incorporating high-quality phrases mined from the corpus as well as tailoring dictionary according to the given corpus, and demonstrate their effectiveness in ablation experiments.

In future, we plan to further investigate the power and potentials of the \our model with \texttt{Tie or Break} scheme in different languages and domains. Also, the proposed framework can be further extended to other sequence labeling tasks, such as noun phrase chunking.
Moreover, going beyond the classical NER setting in this paper, it is interesting to further explore distant supervised methods for the nested and multiple typed entity recognitions in the future.

%% file: 8-ack.tex
%!TEX root = 0_main.tex

\section*{Acknowledgments}
\label{sect:ack}

We would like to thank Yu Zhang from UIUC for providing results of supervised benchmark methods on the BC5CDR and NCBI datasets.
We also appreciate all reviewers for their constructive comments.
Research was sponsored in part by U.S.\ Army Research Lab. under Cooperative Agreement No.\ W911NF-09-2-0053 (NSCTA), DARPA under Agreement No.\ W911NF-17-C-0099, National Science Foundation IIS 16-18481, IIS 17-04532, and IIS-17-41317, DTRA HDTRA11810026, Google Ph.D.\ Fellowship and grant 1U54GM114838 awarded by NIGMS through funds provided by the trans-NIH Big Data to Knowledge (BD2K) initiative (www.bd2k.nih.gov). Any opinions, findings, and conclusions or recommendations expressed in this document are those of the author(s) and should not be interpreted as the views of any U.S.\ Government. The U.S.\ Government is authorized to reproduce and distribute reprints for Government purposes notwithstanding any copyright notation hereon.